\documentclass[review]{elsarticle}

\usepackage{lineno,hyperref}
\modulolinenumbers[5]

\usepackage{subcaption}
\usepackage{amsfonts}       
\usepackage{booktabs}       
\usepackage{xcolor}
\usepackage{amsmath}

\usepackage{float}

\journal{ArXiv}

\usepackage{numcompress}\biboptions{authoryear}

\begin{document}

\begin{frontmatter}
\title{Regularizing Transformers With Deep Probabilistic Layers}




\author[mymainaddress]{Aurora Cobo Aguilera\corref{mycorrespondingauthor}}
\cortext[mycorrespondingauthor]{Corresponding author}
\ead{acobo@tsc.uc3m.es}

\author[mymainaddress]{Pablo Martínez Olmos}
\ead{pamartin@ing.uc3m.es}
\author[mymainaddress]{Antonio Artés-Rodríguez}
\ead{aartes@ing.uc3m.es}
\author[secondaryaddress]{Fernando Pérez-Cruz}
\ead{fernando.perezcruz@sdsc.ethz.ch}

\address[mymainaddress]{Department of Signal Theory and Communications\\
	Universidad Carlos III de Madrid\\
	Avda. de la Universidad 30, 28911, Leganés, Madrid, Spain}
\address[secondaryaddress]{Swiss Data Science Institute (ETHZ/EPFL)\\
	Universitatstrasse 25, 8006, Zurich, Switzerland}

\begin{abstract}
Language models (LM) have grown with non-stop in the last decade, from sequence-to-sequence architectures to the state-of-the-art and utter attention-based Transformers. In this work, we demonstrate how the inclusion of deep generative models within BERT can bring more versatile models, able to impute missing/noisy words with richer text or even improve BLEU score. More precisely, we use a Gaussian Mixture Variational Autoencoder (GMVAE) as a regularizer layer and prove its effectiveness not only in Transformers but also in the most relevant encoder-decoder based LM, seq2seq with and without attention. 
\end{abstract}


\end{frontmatter}


\section{Introduction}
\label{sec:intro}

Deep Generative Models (DGMs) have become a cornerstone in modern machine learning due to their ability to learn abstract features from high-dimensional spaces to generate new data (\citealp{goodfellow2014generative}, \citealp{kingma2014auto}). In the field of Natural Language Understanding (NLU), state-of-the-art is dominated by attention-based probabilistic models, a class of  explicit DGMs that can be trained with Maximum Likelihood Estimation (MLE) approaches \citep{caccia2018language}.  

Regarding other well known DGMs such as Generative Adversarial Networks or GANs \citep{goodfellow2014generative}, so far for NLU they  have not shown the same outstanding results that they achieve for image processing
(\citealp{zhang2019self}, \citealp{radford2015unsupervised}), mostly due to the discrete nature of the data, which leads to non-differentiable issues, mode collapse and optimization instability (\citealp{lu2018neural}, \citealp{caccia2018language}). To tackle these and other issues, recent contributions propose the use of Reinforcement Learning techniques to optimize the GAN loss function (\citealp{yu2017seqgan}, \citealp{fedus2018maskgan}, \citealp{guo2018long}, \citealp{de2019training}), continuous approximations to discrete sampling (\citealp{jang2017categorical}, \citealp{zhang2017adversarial}), or learning a low-dimensional representation through autoencoders (\citealp{zhao2018adversarially}, \citealp{subramanian2018towards}, \citealp{donahue2018adversarial}, \citealp{yu2018learning}, \citealp{haidar2019latent}, \citealp{haidar2019textkd}, \citealp{rashid2019bilingual}). Besides, explicit DGMs such as variational autoencoders (VAEs) have also been proposed in several NLU approaches again with limited success (\citealp{pagnoni2018conditional}, \citealp{shen2018deconvolutional}, \citealp{gupta2018deep}, \citealp{yang2017improved}, \citealp{shi2019fixing}). Some of the pioneers in this field were \citet{bowman2016generating}, who proposes a RNN-based VAE for text generation. Even in an extent, \citet{hu2017toward} combine a VAE with a discriminator to build a hybrid model that solves the text generation problem. In all these works, both GANs and VAEs are at the core of the NLU model, and hence are fully responsible to capture the semantic structure and generate text. For this particular task, they are still not competitive with attention-based probabilistic models \citep{caccia2018language}.

In this work, we propose to exploit DGMs for NLU in a completely novel and different way. Instead of training a DGM to solve a NLU task, we rely on a hybrid model in which a transformer-based architecture like BERT \citep{devlin2018bert} is combined with a VAE, which is placed inside its structure as a stochastic layer that helps to learn a richer hidden space, enforcing a regularization effect to some extent. In particular, we use a structured VAE that implements a mixture of Gaussians in the latent space (GMVAE) \citep{dilokthanakul2016deep}, since it is able to capture more complex data in an easier way than the traditional vanilla VAE.  In a similar way, \citet{sriram2018cold} and \citet{gulcehre2015using} built fusion models taking advantage of a pre-training process as we explain later. Nevertheless, they only focused on a basic seq2seq architecture.

Regularization in deep learning has risen up from the beginning of Neural Networks with the extensively use of tools such as dropout \citep{srivastava2014dropout}, early stopping, data augmentation or weight decay \citep{krogh1992simple}, which helps models to generalize. However, regularization in NLUs is a much-less explored field and none of these tools experience the same versatility as our proposal in this paper, in which the GMVAE performs a controlled and structured noise injection within the NLU deep network. When combined with BERT, we name our model as NoRBERT (Noisy Regularized BERT) and we conclude that the effect of the stochastic layer is very different depending on the transformer layer where it is placed. If the layer is placed at the end of the structure, it drives more versatile topics when imputing missing words. On the contrary, when placed at the bottom, it improves BLEU score, what coincides with the goal of traditional regularization mechanisms.

We illustrate our approach in word imputation problems (masking the source text corpora) using  a BERT transformer network, demonstrating gains in machine translation setups (better BLUE scores) and the versatility of the method to impute missing words by a large set of examples. Furthermore, we also explore the GMVAE regularization effect in traditional seq2seq models with and without attention mechanisms and explain the regularization functionality in a simple well-known problem as it is classification of Fashion MNIST images. The code to generate our results is available in an open repository\footnote{https://github.com/AuroraCoboAguilera/NoRClassifier}.

This paper is organized as follows. Firstly, in Section \ref{sec:related} we describe some related work which is key to understand the paper: the VAE, and more precisely the GMVAE, as the main structure of the regularizer and transformer networks with BERT as our model to be studied. Secondly, in Section \ref{sec:classifier} we explore a basic example of applying our idea in a well-known scenario as it is Fashion MNIST.  This is a useful prove of the stochastic layer effect and its effectiveness in other problems. Thirdly, in Section \ref{sec:method} we describe in detail our model, NoRBERT, and two variants of it, Top and Deep NoRBERT, depending on the transformer layers where we apply the regularization. Then, we present the results of these two options in Section \ref{sec:experiments}. Moreover, we include an extension (Section \ref{sec:extension}) where we study other relevant encoder-decoder based LMs as it is seq2seq with and without attention \citep{bahdanau2015neural}. Finally, in Section \ref{sec:conclu} we conclude our work and mention some future lines of research.

\section{Related work}
\label{sec:related}

\subsection{Variational Autoencoders with Gaussian mixture priors}
\label{subsec:vae}

\begin{figure}[ht]
	\begin{subfigure}[b]{.49\textwidth}
		\centering
		\includegraphics[width=0.3\linewidth]{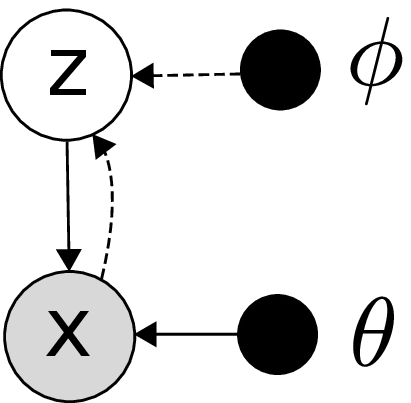}  
		\caption{VAE.}
		\label{fig:graph_vae}
	\end{subfigure}
	\begin{subfigure}[b]{.49\textwidth}
		\centering
		\includegraphics[width=0.55\linewidth]{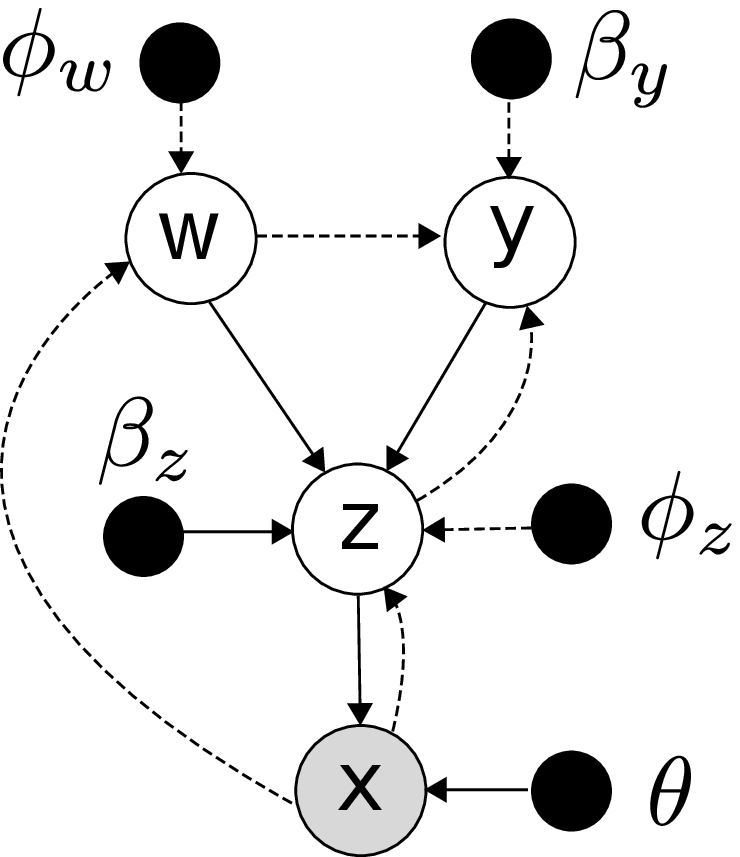}  
		\caption{GMVAE.}
		\label{fig:graph_gmvae}
	\end{subfigure}
	\caption{The directed graphical models into consideration. Solid lines denote the generative model and dashed lines the variational approximation. The shaded variables are considered the observed inputs, the dark units are the networks parameters to be optimized and the units that are left are the latent variables.}
	\label{fig:graph_vae-gmvae}
\end{figure}

A VAE \citep{kingma2014auto} is a class of density estimator that consists on two networks, an encoder and a decoder or generator, that builds a regular latent space with the help of probability distributions. The properties of the organized latent space allow not only the reconstruction of the input data but also the generation of new instances from a sampling procedure. In a standard vanilla VAE, see Figure \ref{fig:graph_vae}, the low-dimensional latent space follows a Gaussian prior distribution likelihood parameters, e.g. mean and covariance matrix of $p(x|z)$ are parameterized with the decoder network with input $x$. Variational inference of the model parameters is achieved by maximizing a lower bound on $\log p(x)$, which in turn depends on a flexible NN parameterized distribution $q(z|x)$ that approximates the true posterior $p(z|x)$:
\begin{equation}
	\label{eq:vae}
	\mathcal{L}_{ELBO} \left( \theta, \phi, x\right)  =  
	\mathbb{E}_{z\sim q_{\phi}\left( z|x\right) } \left[ \log p_{\theta}\left( x|z\right) \right] - 
	\mathcal{K}\mathcal{L} \left[ q_{\phi}\left( z|x\right) \|p\left( z\right) \right],
\end{equation}
where $\mathcal{K}\mathcal{L}(q|p)$ is the KL divergence between distributions $q$ and $p$ and acts as a regularization in the evidence lower bound (ELBO) objective. The graphical model of $q(z|x)$ is indicated in Figure \ref{fig:graph_vae} with dotted lines.

The flexibility of VAEs have encouraged the study of different priors and architectures to obtain models capable of inferring more complex structured data. That is the case of using a Mixture of Gaussians (MoG) as the prior distribution $p(z)$ for the latent space because it helps to capture the multimodal nature of some data (\citealp{jiang2017variational},  \citealp{dilokthanakul2016deep}). We refer to this model as GMVAE, and its graphical model is shown in Figure \ref{fig:graph_gmvae}. The GMVAE generative model proposed by \citet{dilokthanakul2016deep} is characterized by the following distributions:
\begin{subequations}
	\begin{align}
		p(z) & = \int p(z|w, y) \cdot p(w) \cdot p(y) dw dy \label{eq:p1} \\
		p(w) & = \mathcal{N} (0, \mathcal{I}) \label{eq:p2} \\
		p(y) & = \text{Mult}(\pi), \quad \pi_i = \frac{1}{K} \label{eq:p3} \\
		p_{{\beta}_z}\left( z |w, y\right)  &= 
		\quad \prod_{k=1}^{K} \mathcal{N} ( \mu_{\beta_{y_k}}(w), \Sigma_{\beta_{y_k}}(w))^{y_k == 1} \label{eq:p4} \\
		p_\theta(x|z)  &= \mathcal{N} (\mu_\theta(z), \sigma \mathcal{I}), \label{eq:p5}
	\end{align}
\end{subequations}
where $\mu_{\beta_{y_k}}$, $\Sigma_{\beta_{y_k}}$ and $\mu_\theta$ are neural networks. $\mu_{\beta_{y_k}}$ and $\Sigma_{\beta_{y_k}}$ indicate a different NN per component in the mixture of Gaussians and $K$ is the total number of components. The posterior distribution of $z,w$ and $y$ given $x$ is chosen according to the following factorization
\begin{subequations}
	\begin{align}
		&q_{{\phi}_z} (z|x)  =\mathcal{N}(\mu_{{\phi}_z}(x), \Sigma_{\phi_z}(x)) \label{eq:q1} \\
		&q_{{\phi}_w} (w|x)  = \mathcal{N}(\mu_{{\phi}_w}(x), \Sigma_{\phi_w}(x)) \label{eq:q2} \\
		&q_{{\beta}_y} (y_j ==1|w, z)  = 
		\quad \frac{p(y_j ==1) \cdot p_{\beta_z}(z|y_j =1, w)}{\sum_{k=1}^K p(y_k ==1) \cdot p_{\beta_z}(z|y_k =1, w)}, \label{eq:q3} 
	\end{align}
\end{subequations}
where again $\mu_{{\phi}_z}$, $\Sigma_{\phi_z}$, $\mu_{{\phi}_w}$, and $\Sigma_{\phi_w}$ are dense neural networks, resulting in the following evidence lower bound (ELBO):

\begin{equation}
	\label{eq:gmvae}
	\begin{split}
		&\mathcal{L}_{ELBO} \left( \theta, \phi, x\right)  = \quad \\
		& \quad \mathbb{E}_{z\sim q_{{\phi}_z} } \left[ \log p_{\theta}\left( x|z\right) \right] - 
		\mathbb{E}_{w\sim q_{{\phi}_w}, \;  y\sim p_{{\beta}_y}} \left[\mathcal{K}\mathcal{L} \left[ q_{{\phi}_z}\left( z|x\right) \|p_{{\beta}_z}\left( z |w, y\right) \right] \right] - \\
		& \quad \mathbb{E}_{z\sim q_{{\phi}_z}, \; w\sim q_{{\phi}_w}} \left[\mathcal{K}\mathcal{L} \left[ p_{{\beta}_y}\left( y|w, z\right) \| p\left( y\right) \right] \right] - 
		\mathcal{K}\mathcal{L} \left[ q_{{\phi}_w}\left( w|x\right) \|p\left( w\right) \right]  
	\end{split}
\end{equation}



\subsection{Transformer networks: BERT}

Over the last couple of years, Transformers \citep{vaswani2017attention} have become a revolution in the field of NLU (\citealp{dai2019transformer}, \citealp{keskar2019ctrl}, \citealp{ma2019tensorized}, \citealp{gu2019levenshtein}, \citealp{yang2020localness}) due to their ability to capture longer-range linguistic structure. Unlike previous works (\citealp{sutskever2014sequence}, \citealp{bahdanau2015neural}, \citealp{luong2015effective}), they rely entirely on self-attention to compute the latent representations of the sentences. 

Transformer-based models are usually applied in a transfer learning perspective (\citealp{devlin2018bert}, \citealp{radford2018improving}, \citealp{song2019mass}, \citealp{radford2019language}, \citealp{yang2019xlnet}, \citealp{lample2019cross}, \citealp{dong2019unified}, \citealp{sun2019ernie}, \citealp{xiao2020hungarian}, \citealp{yang2020hierarchical}) that allows users to train smaller datasets in a specific task quicker and more accurate than doing it from scratch. Firstly, you need a pre-trained model that has learned contextualized text representations in a general unsupervised scenario with a large text corpus. Afterwards, you can fine-tune the model using a small database with the addition of few parameters or layers in a downstream task. This is the case of BERT \citep{devlin2018bert}, which stands out above all, providing a pre-trained Transformer text encoder as a general LM for any downstream task. Since its appearance, several BERT-based models have emerged (\citealp{liu2019roberta}, \citealp{sanh2019distilbert}) and today they dominate the leaderboard\footnote{\url{https://gluebenchmark.com/leaderboard.}} in GLUE benchmarks \citep{wang2019glue}.

\begin{figure}[ht]
	\centering
	\includegraphics[width=0.5\linewidth]{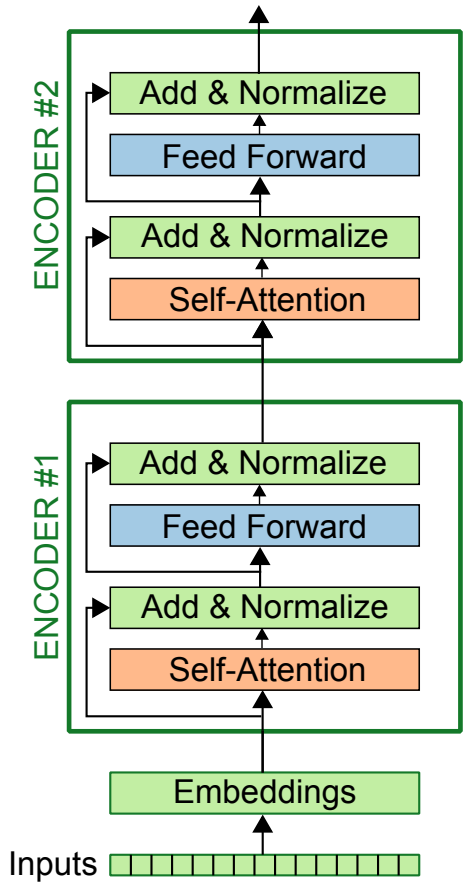}
	\caption{Diagram of BERT structure.}
	\label{fig:BERT}
\end{figure}

Figure \ref{fig:BERT} shows a diagram with the structure of BERT for the first two layers. Basically, it is composed of a first step with the computation of the input sentences embeddings, then a pile of transformer encoder layers, and then, if it is necessary we can apply any task-specific layer on top. Each of these layers consists on two blocks, a multi-head self-attention mechanism and a feed forward network, both with a normalization following them. Regarding the implicit regularization mechanisms within BERT, dropout and weight decay are applied through all the structure: in the fully connected layers in the embeddings, encoder, pooler and in the attention probabilities with rates of $0.1$ a $0.01$ respectively.

BERT makes use of WordPiece embbedings \citep{wu2016google} with a vocabulary size of $30000$ tokens. The base models are pre-trained in the datasets of Book Corpus \citep{zhu2015aligning} with $800$M words and English Wikipedia with $2500$M words.

Although we focus our work in NoRBERT, we will extend the results to traditional seq2seq models (Section \ref{sec:extension}) in order to explain how our mechanism works and show its ability to be integrated in other architectures.

\section{GMVAE as a regularizer in deep neural networks}
\label{sec:classifier}

In this work, we put forward GMVAEs as a robust stochastic layer to enforce regularization in a deep NN, with particular focus on Transformers and NLU. Before describing the methodology in a complex transformer based network, we want to illustrate our approach in a simpler setup, in which we regularize a deep six-layer MLP over the Fashion MNIST (FMNIST) database\footnote{\url{https://github.com/zalandoresearch/fashion-mnist}} \citep{xiao2017fashion}. 

\begin{figure}[ht]
	\centering
	\includegraphics[width=0.5\linewidth]{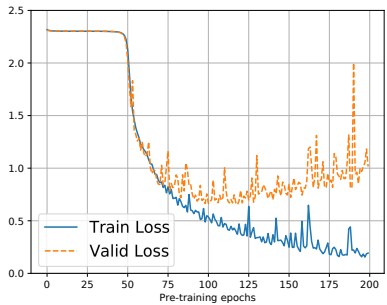}
	\caption{Pre-training a six layer MLP for FMNIST.}
	\label{fig:pretraining_FMNIST}
\end{figure}

In Figure \ref{fig:pretraining_FMNIST} we show the train/validation cross entropy loss of the NN in a completely unregularized training (no dropout or weight decay whatsoever). Validation error begins to raise up from epoch 120. Now we perform the following experiment. We get the NN parameters at epoch 119, at which overfitting was not yet noticeable, and we introduce two types of regularization layers between the first two MLP layers:
\begin{enumerate}
    \item A standard dropout layer with erase probability $p$. 
    \item A GMVAE layer trained using the 700-dimensional internal representation of the first MLP layer. For every output from the first MLP layer, the GMVAE layer first computes a latent low-dimensional representation sampling from the GMVAE posterior distribution in \eqref{eq:q1}-\eqref{eq:q3} to then provide at the output a reconstruction sampled from generative model in \eqref{eq:p1}-\eqref{eq:p5}.  
\end{enumerate}

\begin{figure}[ht]
	\centering
	\includegraphics[width=0.5\linewidth]{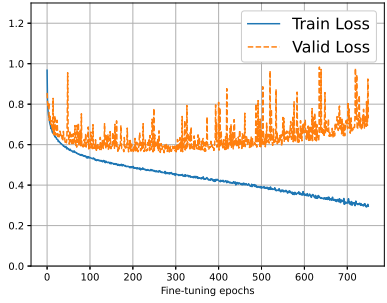}
	\caption{Fine-tuning a six layer MLP for FMNIST with the GMVAE regularization layer placed after the first MLP layer.}
	\label{fig:finetune_FMNIST2}
\end{figure}

\begin{figure*}[ht]
	\begin{subfigure}[b]{.49\textwidth}
		\centering
		\includegraphics[width=.99\linewidth]{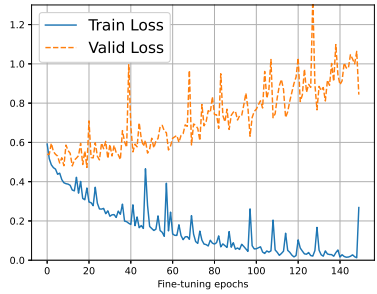}  
		\caption{With dropout probability of $0.1$.}
		\label{fig:FMNIST_dropout1}
	\end{subfigure} %
	\begin{subfigure}[b]{0.49\textwidth}
		\centering
		\includegraphics[width=.99\linewidth]{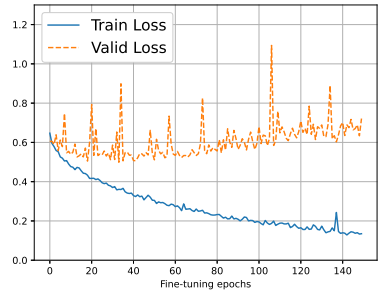}  
		\caption{With dropout probability of $0.5$.}
		\label{fig:FMNIST_dropout}
	\end{subfigure}
	\caption{Fine-tuning a six layer MLP for FMNIST with dropout in the first layer.}
	\label{fig:finetune_FMNIST1}
\end{figure*}

The details about the GMVAE layer parameters used for this experiment can be found in \ref{subsec:appendixConf_fmnist}. Note that the GMVAE layer, as dropout, is introducing a certain level of distortion over the input vector but, unlike dropout, such distortion is not independent to the input vector, as for some atypical vectors the reconstruction noise will be larger. This allows the network to explore diverse regions at the input of the following layer. In Figure \ref{fig:finetune_FMNIST2} we show the train/validation cross entropy loss when layer 1 is fixed (so the GMVAE input distribution is not changing) and we keep training MLP layers 2-6. In Figure \ref{fig:finetune_FMNIST1} we show the performance when dropout with $p=0.1$ (a) and $p=0.5$ (b) is used instead of the GMVAE layer.
On the one hand, observe the inability of dropout to compensate the overfitting of the network. On the other hand, due to the controlled noise injection, the GMVAE avoids overfitting even after an excess of additional epochs. With these figures we can state that the training loss decays much more slowly in our model with a score of $0.3$ after $700$ epochs, while in the dropout case it drops off almost to zero after $150$ epochs.

With this example, we simply want to put forward the use of a DGM (a GMVAE in our case) as potential regularizer with additional flexibility, compared to simpler solutions such as dropout. A detailed cross-validation analysis of what kind of regularization method optimizes the classification performance in this particular classification setting is not relevant at this point. In the following, we show how the use of GMVAE layers if able to enhance the performance of complex pre-trained networks such as BERT, which of course has already been trained with its own regularization methods (including dropout).

\section{Improving BERT with GMVAE layers: NoRBERT}
\label{sec:method}

\subsection{Overview}
The main idea of our work is the integration of the GMVAE in BERT through NoRBERT. In this hybrid model, the GMVAE layer alters the BERT hidden embeddings in one particular layer through a  project-and-reconstruct operation, adding a structured noise to them and hence enforcing a regularization mechanism. In other words, we try to break the determinism in exchange of more robust solutions.  Unlike in other regularization techniques such as dropout, the reconstruction error plus the observation noise (GMVAE noise for short) of the GMVAE will not be uniform across embeddings, since atypical embeddings will suffer from larger GMVAE noise variance. As a result, the network training will rely less on such noisy embeddings, which we show is beneficial for the overall performance.

We want to stress the fact that we use BERT as an exemplary case of how a certain neural language model can be enhanced by the inclusion of GMVAE layers within. Furthermore, in Section \ref{sec:extension} we show how to incorporate the same idea in seq2seq language models with attention. Moving back to BERT, NoRBERT builds upon a pre-trained BERT model, allowing the integration of the GMVAE in an intermediate step. We follow these four main steps:

\begin{enumerate}
	\item Pre-train BERT with a masked text corpora. 
	\item Train a GMVAE over the space of hidden embeddings coming from input sentences using one particular BERT layer. 
	\item Include the GMVAE layer inside the structure. The GMVAE will be responsible for adding noise in the propagation of the information, as in the GMVAE layer every input vector is projected into a low-dimensional space and reconstructed back by sampling from the generative model. 
	\item Retrain the model by fine-tuning all layers above the GMVAE one. The layers below the GMVAE one are not altered so we do not modify the embedding space in which the GMVAE was trained on.
\end{enumerate}

Regarding the base BERT model, for the implementation we use the base model from \citet{devlin2018bert}. In the training we use the masked language modeling (MLM) strategy as \citet{liu2019roberta}, since it is the straightforward strategy to train transformers in word imputation \citep{song2019mass}. 

\subsection{Top and Deep NoRBERT}
\label{subsec:method_NoRBERT}

In the study of NoRBERT we explore placing the regularizer in different layers from BERT. Firstly, we show the effect of the GMVAE on top of the transformer encoder, just before the classification layer that computes the vocabulary logits. We refer to this case as \emph{Top NoRBERT}. Secondly, we explore the consequences when the biggest part of BERT is retrained after placing the GMVAE in one of the first and middle layers. This is referred to \emph{Deep NoRBERT}.

\begin{figure*}[ht]
	\begin{subfigure}[b]{.49\textwidth}
		\centering
		\includegraphics[width=.8\linewidth]{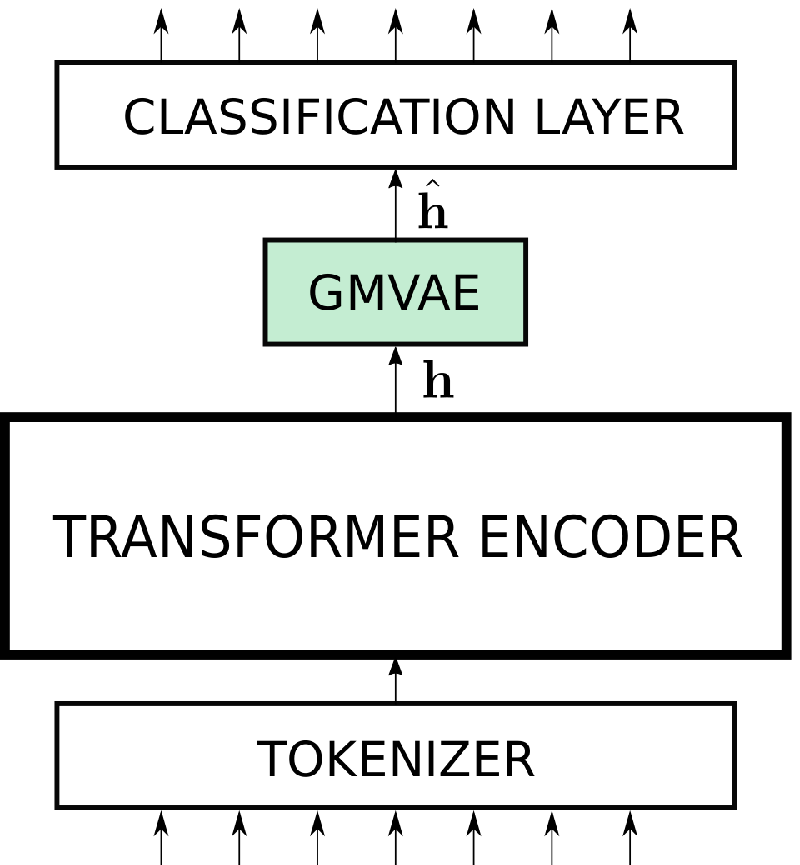}  
	\end{subfigure} %
	\begin{subfigure}[b]{0.49\textwidth}
		\centering
		\includegraphics[width=.8\linewidth]{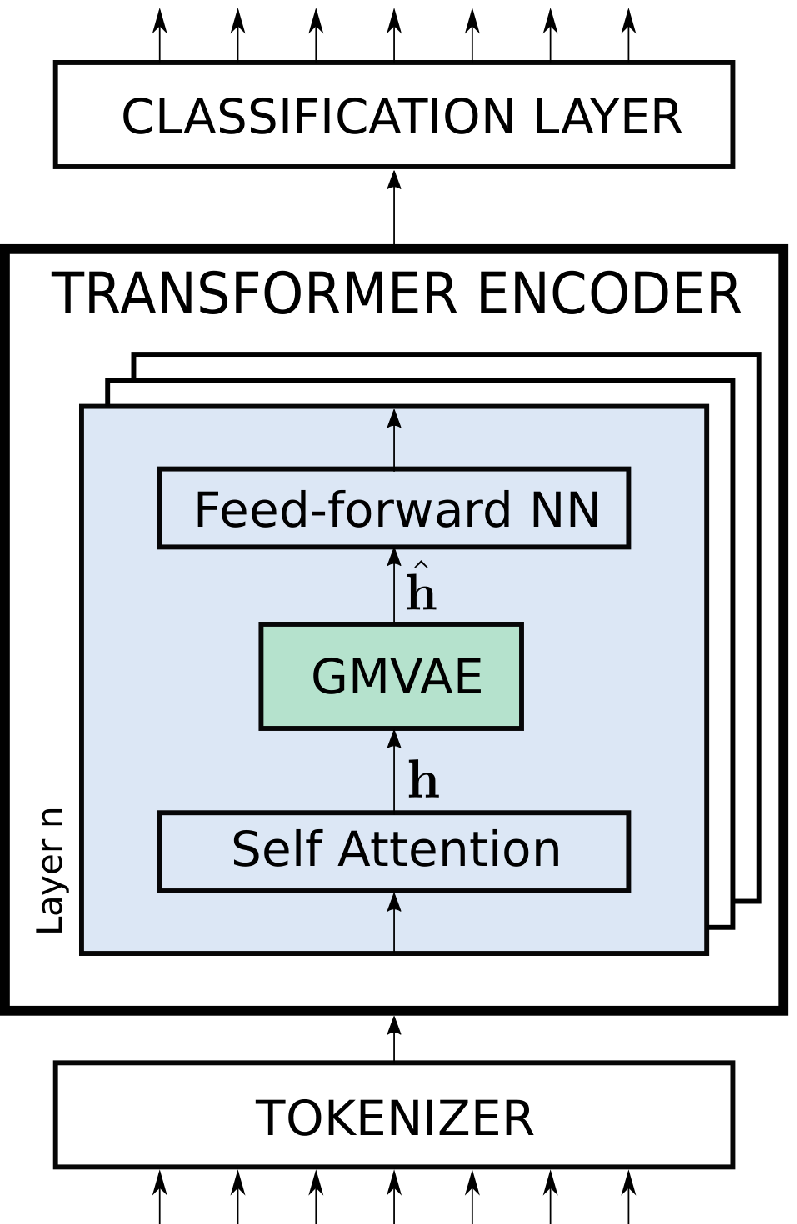}  
	\end{subfigure}
	\caption{Top NoRBERT (a) and Deep NoRBERT (b).}
	\label{fig:NoRBERT}
\end{figure*}

Top NoRBERT consists on a new version of BERT, with a stochastic layer on the top of the transformer encoder as represented in Figure \ref{fig:NoRBERT}(a). Therefore, the only difference from the original model is that we use a GMVAE to reconstruct the last hidden states before the final token decision. We previously train the GMVAE with the hidden states computed by base BERT for the training sentences. Afterwards, we fine-tune the classification layer of BERT with the stochastic reconstruction integrated.

In Deep NoRBERT we include the GMVAE stochastic layer inside an intermediate transformer layer, after the self-attention and before the feed-forward blocks as shown in Figure \ref{fig:NoRBERT}(b).  We fine-tune the parameters in the structure above the regularizer, that is, the feed-forward block in the same encoder layer and the transformer layers that are on top of it. In our experimental results, we demonstrate gains w.r.t. the base BERT model by including only one GMVAE layer. We explore the method in several layers with results that are qualitatively very different compared to Top NoRBERT.

\section{Results}
\label{sec:experiments}

To implement NoRBERT, we make use of the pre-trained base model from BERT described by \citet{devlin2018bert}. This version of BERT is composed of 12 layers, a hidden size of 768 and 12 heads and we make use of the parameters eased by the \textit{Hugging face} library\footnote{https://huggingface.co/} using a MLM objective. We keep the original configuration following the paper \citep{devlin2018bert} except for the hyperparameters mentioned in the following sections. On each experiment, we train a GMVAE using the hidden vectors at some point of BERT structure obtained from training samples with this base model. Once the GMVAE has converged, we build a new architecture based on BERT with the integration of a stochastic layer in the corresponding place of the hidden vectors. This new layer consists on the reconstruction of the hidden vectors through the generative network of the GMVAE. Finally, we fine-tune this new architecture, freezing all parameters below the stochastic layer in the computational graph. 

In the experiments we employ the dataset \textit{snli}\footnote{https://nlp.stanford.edu/projects/snli/} \citep{bowman2015large} with different strategies in the masking process of tokens. It has a vocabulary size of $36711$ different words. We use the entire preprocessed training data which contains $714667$ sentences and a test set of $13350$. The GMVAE is trained for all the tokens of a random set of $50000$ training sentences. When saving the hidden states to train the GMVAE a posteriori, we treat each token as an independent input to the GMVAE, ignoring tokens that correspond to padding (they exist due to BERT format of the tokenizer).

To speed up the loading of data, we utilize the extension \textit{hdf5} for saving the files. Moreover, this way we avoid memory issues when loading the datasets in the programs since these files with the hidden states have significant sizes.

\subsection{Deep NoRBERT}
\label{subsubsec:exp_deepNoRBERT}

First, we present the results of Deep NoRBERT, in which the GMVAE stochastic layer is placed in an intermediate BERT encoder layer, see Section \ref{subsec:method_NoRBERT}. Next we present the results obtained in terms of accuracy and BLEU score for different locations of the GMVAE layer inside the BERT structure. 

The GMVAE layer is trained for $500$ epochs with a learning rate of $5 \cdot {10}^{-5}$. The GMVAE latent  dimension $z$ is set to $150$, the $w$ dimension to $50$, and we consider a mixture of $20$ Gaussians, dropout $0.3$ and networks with a depth of $6$ layers. Then, deep NoRBERT is trained for $8$ epochs freezing the parameters below the stochastic layer. The baseline BERT is also fine-tuned in the same dataset for $8$ epochs so we can make a fair comparison in their performance in missing data imputation. We evaluate the percentage of tokens that are exactly the same as the source sentence in a 1-by-1 comparison. We test two different scenarios, with masked tokens and with disrupted tokens, that is, instead of using the [MASK] token which indicates `unknown', we place random choices from the vocabulary that damage the source sentence. We replicate the random words substituted on each experiment maintaining the same seed in the training. Regarding the masks, 40$\%$ of the sentences chosen at random have at least one [MASK] token, which always replaces a meaningful word (we avoid masks over stopping words).

\begin{table}[h]
	\centering
	\begin{tabular}{crr}
		\hline 
		\textbf{Model} & Masked  & Disrupted \\
		\hline
		BERT & $97.13 \%$  & $96.98 \%$  \\
		1-Deep NoRBERT & $\mathbf{97.32}\boldsymbol{\%}$   & $\mathbf{97.11}\boldsymbol{\%}$ \\
		2-Deep NoRBERT & $\mathbf{97.20}\boldsymbol{\%}$   & $\mathbf{97.07}\boldsymbol{\%}$ \\
		3-Deep NoRBERT & $\mathbf{97.18}\boldsymbol{\%}$   & $\mathbf{97.1}\boldsymbol{\%}$ \\
		9-Deep NoRBERT & $96.87 \%$   & $96.25 \%$  \\
		11-Deep NoRBERT & $96.05 \%$   & $95.34 \%$    \\
		12-Deep NoRBERT & $95.89 \%$  & $93.89 \%$  \\
		\hline
	\end{tabular}
	\caption{Accuracy of different models comparing the unmasked source sentence with the reconstruction. We evaluate a version that keeps the [MASK] tokens and other that substitutes them by random tokens from the vocabulary. In $l$-Deep Norbert, $l$ refers to the transformer BERT layer in which the GMVAE is placed.}
	\label{table:deepNoRBERTResults}
\end{table}

Table \ref{table:deepNoRBERTResults} shows the imputation accuracy for different configurations, in which $l$-Deep NoRBERT means that we placed the GMVAE layer in the $l$-th transformer layer.  For a better visualization, we highlight in bold every case that outperforms the baseline.  Observe that the largest gains are obtained when the GMVAE layer is placed in the bottom of the network, outperforming BERT after fine-tuning. We remark that BERT is a state-of-the-art model for MLU that is pre-trained over a massive dataset and hence any improvement is not negligible, particularly when is achieved by placing a single regularization layer within. Despite some studies about BERT state that the last layers encode task-specific features \citep{kovaleva2019revealing}, our results demonstrate that fine-tuning and regularization of deep layers may improve the overall performance.

\begin{table}[h]
	\centering
	\begin{tabular}{crrr}
		\hline 
		\textbf{Model/Missing rate} & Low  & Medium & High \\
		\hline
		BERT & $86.07$  & $49.43$  & $25.14$ \\
		1-Deep NoRBERT & $\mathbf{86.90}$   & $\mathbf{49.91}$ & $\mathbf{25.53}$ \\
		2-Deep NoRBERT & $\mathbf{86.65}$   & $\mathbf{49.75}$ & $\mathbf{25.26}$ \\
		3-Deep NoRBERT & $\mathbf{86.53}$   & $49.33$ & $\mathbf{25.45}$\\
		9-Deep NoRBERT & $85.52$   & $46.04$ & $21.47$  \\
		11-Deep NoRBERT & $83.89$   & $43.34$ & $19.28$  \\
		12-Deep NoRBERT & $80.77$   & $40.83$ & $17.16$  \\
		\hline
	\end{tabular}
	\caption{BLEU score of different models comparing different missing rates.}
	\label{table:deepNoRBERTResults2}
\end{table}

Table \ref{table:deepNoRBERTResults2} presents the BLEU score obtained by Deep NoRBERT with different layer configurations. We explore different policies of generating missing tokens. `Low' refers to the same mechanism as in Table \ref{table:deepNoRBERTResults} experiments. 
In the policies called `Medium' and `High' we do not exclude any token by its grammatical meaning and mask every word independently with probabilities of $0.4$ and $0.6$ respectively. Table \ref{table:deepNoRBERTResults3} results, called Masked BLEU, differ from the previous ones in the n-grams taken for the metric computation. That is, we only consider n-grams that include a masked token. From both tables we draw similar conclusions: the best performance is obtained when the GMVAE layer is placed at the bottom of the network, right after the first transformer layer.

\begin{table}[h]
	\centering
	\begin{tabular}{crrr}
		\hline 
		\textbf{Model/Missing rate} & Low  & Medium & High \\
		\hline
		BERT & $3.73$  & $21.3$  & $15.34$ \\
		1-Deep NoRBERT & $\mathbf{3.88}$   & $\mathbf{22.7}$ & $\mathbf{16.44}$ \\
		2-Deep NoRBERT & $\mathbf{3.88}$   & $\mathbf{22.50}$ & $\mathbf{16.22}$ \\
		3-Deep NoRBERT & $\mathbf{3.90}$   & $\mathbf{22.28}$ & $\mathbf{16.56}$\\
		9-Deep NoRBERT & $\mathbf{3.87}$   & $19.78$ & $13.34$  \\
		11-Deep NoRBERT & $3.65$   & $18.21$ & $11.64$  \\
		12-Deep NoRBERT & $3.01$   & $16.31$ & $9.61$  \\
		\hline
	\end{tabular}
	\caption{Masked BLEU score of different models comparing different missing rates. }
	\label{table:deepNoRBERTResults3}
\end{table}

\subsection{Top NoRBERT}
\label{subsubsec:exp_topNoRBERT}

The above results demonstrate that retraining BERT when we include a GMVAE layer within may bring imputation improvement when the layer is placed deep inside the BERT network. From this perspective, placing the GMVAE layer in the top of the network, as we do in Top NoRBERT, lacks a priori of any interest. Actually, when we freeze all the parameters from the encoder layers and fine-tune only the classification layer we achieve an imputation accuracy of $77.14 \%$ (Masked)  and $75.53 \%$  (Disrupted), far below the Deep NoRBERT performance in Table \ref{table:deepNoRBERTResults}. A closer look to the actual imputed words by Top NoRBERT in different sentences led us to conclude that the final GMVAE layer placed right below the classifier promotes topic diversity in the imputation task, which would explain the severe drop in accuracy w.r.t. Deep NoRBERT. This result may be consequence of the fact that upper layers in BERT learn specific features that affect the token choice while the deeper layers pick up general characteristics of text. 

Therefore, in order to visualize the effect of our regularizer, Table \ref{table:topNoRBERTResults} includes some test sentences reconstructed by Top NoRBERT in comparison with the baseline BERT. In \ref{sec:appendixE} we have included more examples with longer sentences (Table \ref{table:topNoRBERTResults2}) as an extension. For the generation of the results we use again the \textit{snli} dataset with the masking policy defined as `Low'. The baseline corresponds to BERT model fine-tuned for half an epoch and a learning rate of $5 \cdot {10}^{-5}$. The training of Top NoRBERT was fine-tuned with the same configuration. Regarding the GMVAE, we maintained all the previous parameters, except that we increased the learning rate to ${10}^{-4}$ and trained $200$ epochs.

\begin{table}[h!]
	\centering
	\begin{tabular}{l}
		\toprule
		\textbf{Source:} \underline{This} church \underline{choir} sings \underline{to the masses as they sing joyous songs}\\ \hspace{1cm} \underline{from the book at a church .}  \\
		\textbf{BERT:} this \textcolor{red}{large} choir \textcolor{red}{looks} to the \textcolor{red}{camera} as they sing \textcolor{red}{joy about} songs\\ \hspace{1cm} from the book at a church. \\
		\textbf{Top-NoRBERT:} \textcolor{red}{a dancing band performs} to the \textcolor{red}{friends} as they \textcolor{red}{perform} \\ \hspace{1cm} \textcolor{red}{funcy bands} from the book at a \textcolor{red}{museum}.  \\
		\midrule
		\textbf{Source:} \underline{A man} reads \underline{the} paper\underline{ in a bar with green lighting .} \\
		\textbf{BERT:} a man \textcolor{red}{in} the \textcolor{red}{drink} in a bar with green lighting. \\
		\textbf{Top-NoRBERT:} a man \textcolor{red}{on} the \textcolor{red}{bike} in a bar with green \textcolor{red}{lights}.  \\
		\midrule
		\textbf{Source:} \underline{During calf} roping \underline{a cowboy calls off his} horse . \\
		\textbf{BERT:} \textcolor{red}{a the race} a cowboy \textcolor{red}{call} off his \textcolor{red}{back}. \\
		\textbf{Top-NoRBERT:} during \textcolor{red}{horse jumping} a cowboy \textcolor{red}{tries} off his \textcolor{red}{dog}.  \\
		\midrule
		\textbf{Source:} \underline{A man in a black} shirt \underline{is looking at a} bike\underline{ in a workshop .}  \\
		\textbf{BERT:} a man in a black shirt is looking at a \textcolor{red}{woman} in a \textcolor{red}{conference}. \\
		\textbf{Top-NoRBERT:} a man in a black shirt is looking at a \textcolor{red}{sign} in a \textcolor{red}{shop}.  \\
		\midrule
		\textbf{Source:} \underline{The man in the black wetsuit is} walking \underline{out of the water .} \\
		\textbf{BERT:} the man in the black wetsuit is \textcolor{red}{coming} out of the water. \\
		\textbf{Top-NoRBERT:} the man in the black \textcolor{red}{swimsuit} is \textcolor{red}{jumping into} of the \\ \hspace{1cm} water  \\
		\midrule
		\textbf{Source:} \underline{Five girls and two guys are crossing a overpass .}\\
		\textbf{BERT:} Five girls and two guys are crossing a overpass . \\
		\textbf{Top-NoRBERT:} \textcolor{red}{three} girls and two guys are \textcolor{red}{down} a \textcolor{red}{intersection} \\ \hspace{1cm} \textcolor{red}{sidewalk}.  \\
		\bottomrule
	\end{tabular}
	\caption{Examples of sentences reconstructed by Top NoRBERT. The first sentence is the original one, with the observed words underlined, i.e. \textbf{no underlying means a missing word}. The second is the output of the baseline, BERT fine-tuned. Finally, we show our reconstruction. The words in red correspond to mismatches with the original sentence.}
	\label{table:topNoRBERTResults}
\end{table}

As it is shown in Table \ref{table:topNoRBERTResults}, the GMVAE stochastic layer at the top of BERT helps it to reconstruct sentences from a robust space, inducing the generation of more diverse sequences than the baseline. It is interesting how it changes some words maintaining the original structure as in the first example in Table \ref{table:topNoRBERTResults}. Moreover, these alterations maintain grammatical rules (`performs' and `perform' are used according to the subject) and sometimes correspond to synonymous or analogous words (in this same example, the verb `sing' is replaced by `perform', the noun `choir' by `band', the object `masses' by `friends' and the place `choir' by `museum'). This diversity skill is not obtained by the baseline, so it is a characteristic uniquely from our methodology. In other cases, we get changes in words that are not masked so the overall sentence makes sense. The fifth example changes `out' by `into' as a consequence of infering `jumping' from the masked word `walking'. In the last example, NoRBERT changes `crossing a overpass' by `down a intersection sidewalk' as a semantically related structure that also corresponds the verb `to be'.

Enhancing diversity in text generation is a little explored area, as we do not even dispose of clear metrics to measure such an ability, in opposition to for instance image generation, in which researches typically rely on feature space metrics such as the FID to evaluate generation diversity \citep{heusel2017gans}. We believe that the Top NoRBERT strategy to achieve such diversity may open future research lines on this topic.

\section{Extension}
\label{sec:extension}

Finally, we show how the imputation diversity of traditional seq2seq-type models can be also enhanced by including a regularizer GMVAE layer inside their structure. We start with a simple seq2seq model  and then a seq2seq model with attention \citep{luong2015effective}.

\subsection{Seq2seq}
\label{subsec:method_seq2seq}

\begin{figure}[ht]
	\centering
	\includegraphics[width=0.5\linewidth]{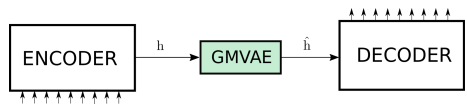}
	\caption{Diagram of the regularized seq2seq model.}
	\label{fig:seq2seq}
\end{figure}

A  seq2seq model is composed of an encoder which maps the input sentence into a fixed-size vector and a decoder to map this vector into a target sentence. In this architecture, we propose to train a GMVAE over the encoder output as shown in Figure \ref{fig:seq2seq}. In the fine-tuning step, the encoder is fixed and the decoder is re-trained taking as inputs the GMVAE noisy reconstructed vectors.

\subsubsection{Results}
\label{subsec:exp_seq}
One of the problems of this first model is caused by the limitations of our baseline. Seq2seq is not suitable for dealing with complex and realistic datasets, that is, long sentences and a wide dictionary, since they encode the semantic and syntactic information of a whole sentence in a single vector, e.g. the encoder output. Notwithstanding, we present in this section some examples where the effect of the regularization layer can be evaluated. The configuration details can be consulted in \ref{subsec:appendixConf_seq2seq}. 

In Table \ref{table:seq2seqResults} we show some test sentences reconstructed by our model, compared with the baseline, which is the pre-trained seq2seq model without any GMVAE stochastic layer. In both cases, we reconstruct with the most likely word. By its own, a seq2seq model fulfills its task if the dataset is not very complex, so we have restricted the results to that premise (refer to \ref{sec:appendixDataset} for dataset details). Our model achieves its goal when the sentences are short, finding words that fit the holes, while the baseline fails more often in the task, repeating the previous or following word into the masked place when it does not predict anything better (examples 2, 3 and 5). In addition, our method is able to change other words in the sentence, even if they were not masked, so the overall construction has more sense (example 2). However, in complex scenarios (example 5), both tend to fail, above all our approach, with not grammatically correct sentences.

\begin{table*}[h!]
	\centering
	\begin{tabular}{l}
		\toprule
		\underline{a woman standing in a dark}  doorway , \underline{waiting to be let into the building .}  \\
		\underline{a woman standing in a dark}  \textcolor{red}{small game} \underline{waiting to be let into the building .} \\
		\underline{a woman standing in a dark}  \textcolor{red}{blue jacket} \underline{waiting to be let into the building .}  \\
		\midrule
		\underline{a man in an} orange \underline{hat starring at} something .  \\
		\underline{a man in an} \textcolor{red}{hat} \underline{hat starring at} \textcolor{red}{many} .  \\
		\underline{a man in an} \textcolor{red}{orange shirt performs} \underline{at} \textcolor{red}{night} .  \\
		\midrule
		\underline{the red car is} ahead \underline{of the} two \underline{cars in the background .}  \\
		\underline{the red car is} \textcolor{red}{is} \underline{of the} \textcolor{red}{cars} \underline{cars in the background .}  \\
		\underline{the red car is} \textcolor{red}{is} \underline{of the} \textcolor{red}{street} \underline{cars in the background}  \\
		\midrule
		\underline{five} people wearing \underline{winter jackets and helmets stand in the snow , with} \\ \hspace{1cm} snowmobiles \underline{in the background.}  \\
		\underline{five} \textcolor{red}{girls ,} winter jackets and helmets stand in the snow , with \textcolor{red}{flowers} in \\ \hspace{1cm} the background .  \\
		\underline{five} \textcolor{red}{soccer ,} winter \textcolor{red}{teenager} and \textcolor{red}{others} stand \\ \hspace{1cm} in the snow with \textcolor{red}{this river} in the background .  \\
		\midrule
		\underline{a} large \underline{bull targets a man , inches away , in a} rodeo \underline{with his horns , while} \\ \hspace{1cm} \underline{a rodeo clown runs} \ldots  \\
		a \textcolor{red}{bull}  bull targets a man , \textcolor{red}{petting} away , in a \textcolor{red}{bottle} with his \textcolor{red}{other} , while \\ \hspace{1cm}a rodeo clown \textcolor{red}{tries} \ldots  \\
		a young \textcolor{red}{boy move} a \textcolor{red}{shoeshine opponent head} , \textcolor{red}{wearing a blue} with \textcolor{red}{the} \\ \hspace{1cm} \textcolor{red}{girl} , \textcolor{red}{with two boys} \ldots \\
		\bottomrule
	\end{tabular}
	\caption{Examples of sentences reconstructed by the regularized seq2seq. The first sentence is the original one, with the observed words underlined, i.e. \textbf{no underlying means a missing word}. The second is the output of the baseline seq2seq pre-trained. Finally, we show our method. The words in red correspond to mismatches with the original sentence.}
	\label{table:seq2seqResults}
\end{table*}

\subsection{Seq2seq with attention}
\label{subsec:method_seqAtt}

We include now the global attention mechanism into the seq2seq \citep{luong2015effective}, allowing the network to focus on the relevant parts of the source sentence, acting as an alignment system between encoder and decoder and improving the performance. Now, as the decoder attends to the encoder hidden states at each time step, the previous approach (Section \ref{subsec:method_seq2seq}) results to be insufficient. In this section we present two kind of methodologies: the option 1 regularizes the hidden states in the decoder LSTMs with a Conditional GMVAE (C-GMVAE), and the option 2 the attention vectors with a GMVAE.

\begin{figure*}[ht]
	\begin{subfigure}[b]{.49\textwidth}
		\centering
		\includegraphics[width=.95\linewidth]{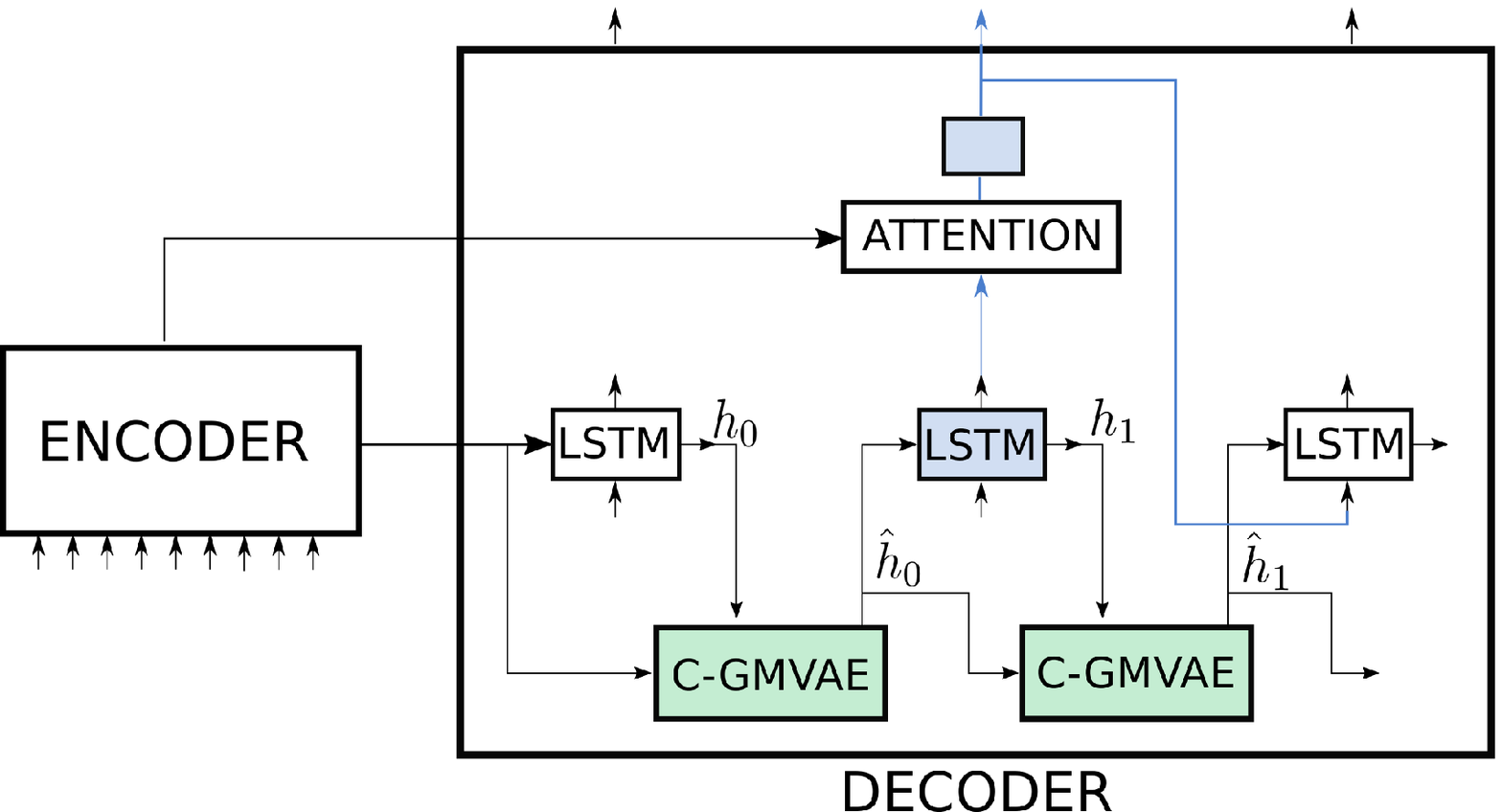}  
		\caption{Option 1.}
		\label{fig:seq2seqAtt1}
	\end{subfigure} %
	\begin{subfigure}[b]{0.49\textwidth}
		\centering
		\includegraphics[width=.95\linewidth]{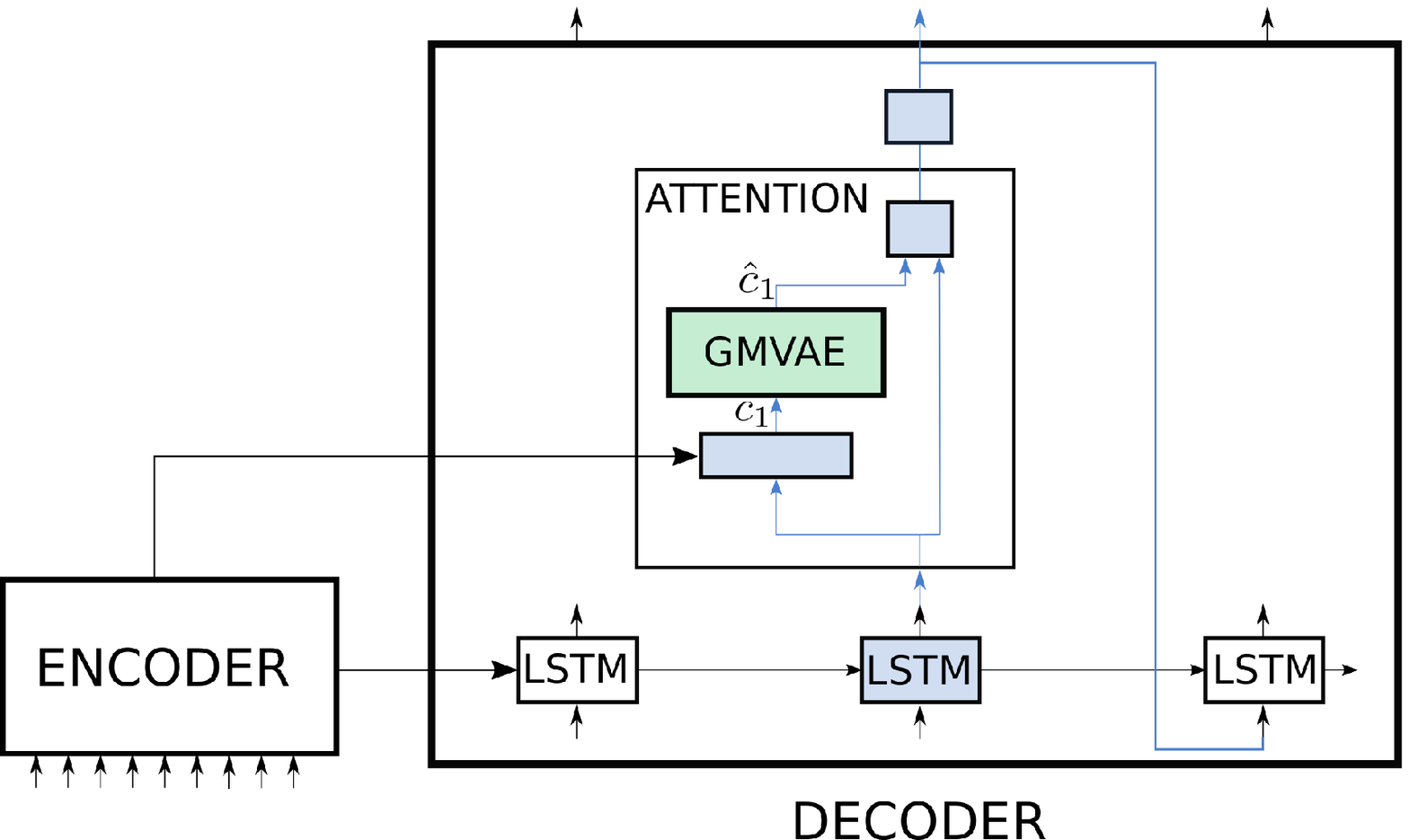}  
		\caption{Option 2.}
		\label{fig:seq2seqAtt2}
	\end{subfigure}
	\caption{Diagrams of the GMVAE regularized seq2seq model with attention.}
	\label{fig:seq2seqAtt}
\end{figure*}

The option 1 aims to regularize the hidden states of the decoder at each step ($h_0, h_1, ... h_T$). To achieve this task, we train a C-GMVAE with pairs of consecutive hidden states ($h_i, h_{i+1}$) from the training sentences, the first one acting as the conditioning input and the second as the input to be reconstructed. See \ref{sec:appendixA} for details on the C-GMVAE.  At each time step, the C-GMVAE receives the previous state and the current hidden state ($\hat{h}_{i-1}, h_{i}$) to reconstruct the latter (${\hat{h}}_{i}$). Figure \ref{fig:seq2seqAtt1} shows a diagram with this approach. We highlight in blue the process concerning the step $i=1$ as an example, but it is repeated from the beginning until the end-of-sentence token is generated.

Option 2 is structurally simpler.  We incorporate the noise in a controlled way, avoiding dependencies on previous states. For that, we propose to introduce the GMVAE layer inside the attention mechanism itself. In particular, the GMVAE layer is trained over the context vectors ($c_0, c_1, ... c_T$).   This model is shown in Figure \ref{fig:seq2seqAtt2}, where we train the GMVAE with the context vectors of words from the training sentences. We treat each token independently in the GMVAE, since the context vectors usually attend to no more than one or two tokens, thus not requiring a conditional GMVAE.

\subsubsection{Results}
\label{subsec:exp_seqAtt}

Table \ref{table:seq2seqAttResults} shows the results of the two configurations proposed. We use the same dataset as previously but as the model is more powerful due to attention, we are able to increase the percentage of masked tokens to be inferred (see \ref{sec:appendixDataset} for details of this second strategy) without damaging the overall performance of the seq2seq. Moreover, in \ref{sec:appendixE} we extend the results for a larger text corpora. The configuration details are presented in \ref{subsec:appendixConf_seq2seqAtt}.

\begin{table}[t]
	\centering
	\begin{tabular}{l}
		\toprule
		a \underline{man} in an orange \underline{hat starring at} something .  \\
		a \underline{man} in a \textcolor{red}{black} \underline{hat starring at} something .  \\
		a \underline{man} in a \textcolor{red}{hard} \underline{hat starring at} something .  \\
		a \underline{man} in a \textcolor{red}{black} \underline{hat starring at} something .  \\
		\midrule
		a boston \underline{terrier} is \underline{running} on lush \underline{green grass} in \underline{front of} a white fence . \\
		a \textcolor{red}{gray} \underline{terrier} \textcolor{red}{dog} \underline{running} on \textcolor{red}{the} \underline{green grass} in \underline{front of } a \textcolor{red}{blue shack} .  \\
		a \textcolor{red}{gray} \underline{terrier} \textcolor{red}{dog} \underline{running} \textcolor{red}{through tall} \underline{green grass } in \underline{front of} a \textcolor{red}{red ball}   \\
		a \textcolor{red}{black dog} is \underline{running} \textcolor{red}{through the grass} \underline{grass} in \underline{front of} a \textcolor{red}{red flag} .  \\
		\midrule
		a girl in karate \underline{uniform} breaking \underline{a stick} with a front \underline{kick .}  \\
		a \textcolor{red}{man in a} \underline{uniform}	\textcolor{red}{ throws} \underline{a stick}	\textcolor{red}{ to his his} \underline{kick .}  \\
		a \textcolor{red}{boy in a} \underline{uniform} 	\textcolor{red}{with} \underline{a stick }	\textcolor{red}{in a large} \underline{kick .}  \\
		a \textcolor{red}{man in a} \underline{uniform} 	\textcolor{red}{kicking} \underline{a }	\textcolor{red}{ball up to his opponent}  .  \\
		\midrule
		five people wearing \underline{winter} jackets and helmets \underline{stand} in the snow , \underline{with} \\ \hspace{1cm} snowmobiles in the \underline{background} .  \\
		\textcolor{red}{two men in }\underline{winter} jackets and \textcolor{red}{hats }stand in \textcolor{red}{a large space} with \textcolor{red}{structure} \\ \hspace{1cm} in the background .  \\
		\textcolor{red}{a group of} \underline{winter} \textcolor{red}{day at a} stand in \textcolor{red}{a snowy area} with \textcolor{red}{trees} in the background  \\
		\textcolor{red}{two men wearing} \underline{winter} \textcolor{red}{clothing} and \textcolor{red}{hats} stand \textcolor{red}{on the snow covered} \\ \hspace{1cm} \textcolor{red}{street} with \textcolor{red}{flags open .} \\
		\midrule
		\underline{a} 	man in a \underline{vest} is \underline{sitting} in a chair \underline{and} holding magazines .  \\
		\underline{a} 	man in a \underline{vest} 	is \underline{sitting} \textcolor{red}{on} a \textcolor{red}{rock} \underline{and} 	\textcolor{red}{looking out }.  \\
		\underline{a} man in a \underline{vest}	is \underline{sitting} \textcolor{red}{on} a \textcolor{red}{sidewalk} \underline{and} 	\textcolor{red}{playing music }. \\
		\underline{a} man \textcolor{red}{wearing} a \underline{vest} is \underline{sitting}	\textcolor{red}{on} a \textcolor{red}{wall} \underline{and} 	\textcolor{red}{smoking a cigarette }. \\
		\midrule
		a \underline{mother} and \underline{her} young son \underline{enjoying} a beautiful \underline{day outside}	 .  \\
		a \underline{mother} and \underline{her} \textcolor{red}{daughter are} \underline{enjoying} a	\textcolor{red}{wedding} day outside .\\
		a \underline{mother} and \underline{her} \textcolor{red}{ child are} \underline{enjoying} a \textcolor{red}{hot} day outside . \\
		a \underline{mother} and \underline{her} \textcolor{red}{children are} \underline{enjoying} a \textcolor{red}{hot} day outside . \\
		\bottomrule
	\end{tabular}
	\caption{Examples of sentences reconstructed by the regularized seq2seq with attention following the same format as Table \ref{table:seq2seqResults}: original, baseline, options 1 and 2.}
	\label{table:seq2seqAttResults}
\end{table}

The results in Table \ref{table:seq2seqAttResults} show how both designs fit our goal, generating new sentences and computing substitutes to the masked tokens that fit the gaps. All of the sentences that are exposed belong to the testing dataset and have been selected randomly.  As opposite as in the first scenario in Section \ref{subsec:exp_seq}, the generation of sentences has improved due to the attention mechanism, so both the baseline and our method perform better the reconstruction of sentences, as was expected. Moreover, the option 2, regularizing the context vectors, not only imputes the masked tokens but also some other tokens in the sentence so the complete structure makes sense. For example, in the second sentence the word `terrier' is removed and `dog' is changed of position. More interesting is the third one, where `kick' and `stick' are deleted but `kicking' appears as a verb form of `kick'. 

To understand the diversity of solutions achieved with our model, we can examine not only the most likely imputed word, but also the top five. We focus in  option 2 for simplicity. For example, in the first sentence, the baseline best options for `orange' correspond to colours, however our method also infers the word `cowboy' in the top 5. In the longest sentence, the forth, we found that even if the final reconstruction was not completely correct (neither in the baseline), our method achieves more varied candidates. In particular, the word `snowmobiles' has the more likely alternatives [`structure', `furniture', `each', `it' and `reflections'] for the baseline while ours are [`flags', `trees', `umbrellas', `people' and `something'], which is a more diverse set that absolutely fits the previous word `with'.

Our results demonstrate that our proposal performs at least as good as the baseline but in many times is capable to improve generalization in the imputation of missing words. Even more, it can be seen as a way of data augmentation in the sense that builds new sentences, acceptable and different from the baseline choices.


\section{Conclusions  and future work}
\label{sec:conclu}

In this work we have proved the successful effect of adding a stochastic GMVAE layer in BERT through NoRBERT. We study the different advantages regarding the layer where it is applied. While Top NoRBERT successes with an increment of diversity as well as an easier way of adaptability to new contexts, Deep NoRBERT responds better in terms of accuracy and BLEU score. In the former case, we propose a novel methodology to generate new structures of text with diverse topics that fit the gaps thanks to the inclusion of controlled noise through a DGM. As a way of reinforcing our idea, we prove the GMVAE effect regularizing a well-studied scenario with FMNIST images.

As an extension, in Section \ref{sec:extension} we present the advantages of the stochastic layer in autoregressive seq2seq models. Despite their limitations with long sentences, our method is able to predict assorted structures upon an extend. Then, we enforced the same idea applying attention and exploring other scenarios that incorporate the regularization at different points of the baseline. In this work, we successfully reconstruct a varied set of topics from the masked source sentences and demonstrate the efficacy of the stochastic layer in finding synonymous or analogous fragments that fit in the gaps.

For now, there is no metric to evaluate robust and varied solutions, since traditional evaluations as BLEU \citep{papineni2002bleu} or ROUGE \citep{lin2004looking} are based in the reconstruction of the original sentence. There is no perfect evaluation metrics testing the text generation because it is difficult to resume all the semantic and syntactic properties that language needs to fulfil \citep{wang2019evaluating}.  Therefore, we let for future work the exploration of metrics or lost functions that allows the LM to generate sentence embeddings with more diversity based on the context.

\section*{Acknowledgements}

This work has been partly supported by Spanish government MCI under grants TEC2017-92552-EXP   and   RTI2018-099655-B-100,   by   Comunidad   de   Madrid   under   grants   IND2017/TIC-7618,IND2018/TIC-9649, IND2020/TIC-17372, and Y2018/TCS-4705, by BBVA Foundation under the Deep-DARWiNproject, and   by   the   European   Union   (FEDER)   and   the   European   Research   Council   (ERC)   through   the European Union's Horizon 2020 research and innovation program under Grant 714161. The work by Aurora Cobo has been aditionally funded by Spanish \textit{Ministerio de Educación, Cultura y Deporte}, grant FPU17/03895.

\bibliography{mybibfile}

\newpage

\appendix

\section{The Conditional GMVAE}
\label{sec:appendixA}

When we deal with conditional DGM, we mean that the entire generative process is conditioned on some extra observed inputs. \citet{sohn2015learning} presented Conditional Variational Autoencoder (CVAE), where the observations modulate the Gaussian prior. In a similar way, we have studied two architectures to condition our distributions on an input that we have defined $h$. In this section, we expose the changes applied and describe the two versions of C-GMVAE that we have explored, referred as models A and B.

\begin{figure}[ht]
	\begin{subfigure}[b]{.5\textwidth}
		\centering
		\includegraphics[width=0.65\linewidth]{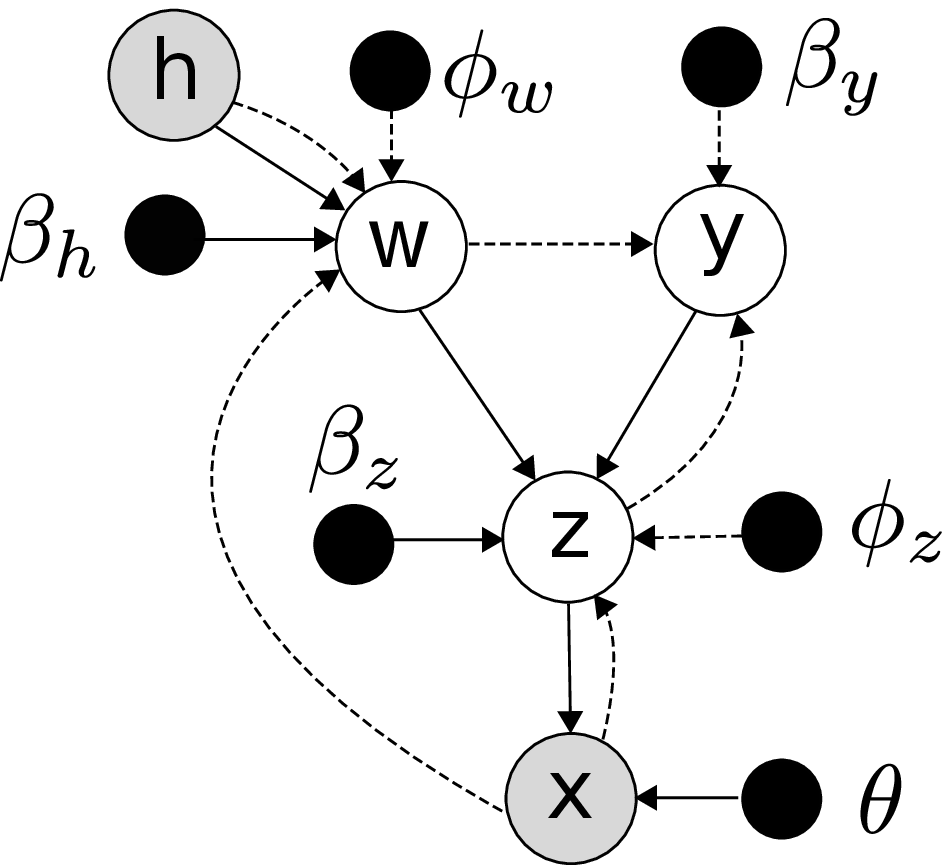}  
		\caption{C-GMVAE model A.}
		\label{fig:graph_cgmvaeA}
	\end{subfigure}
	\begin{subfigure}[b]{.5\textwidth}
		\centering
		\includegraphics[width=.5\linewidth]{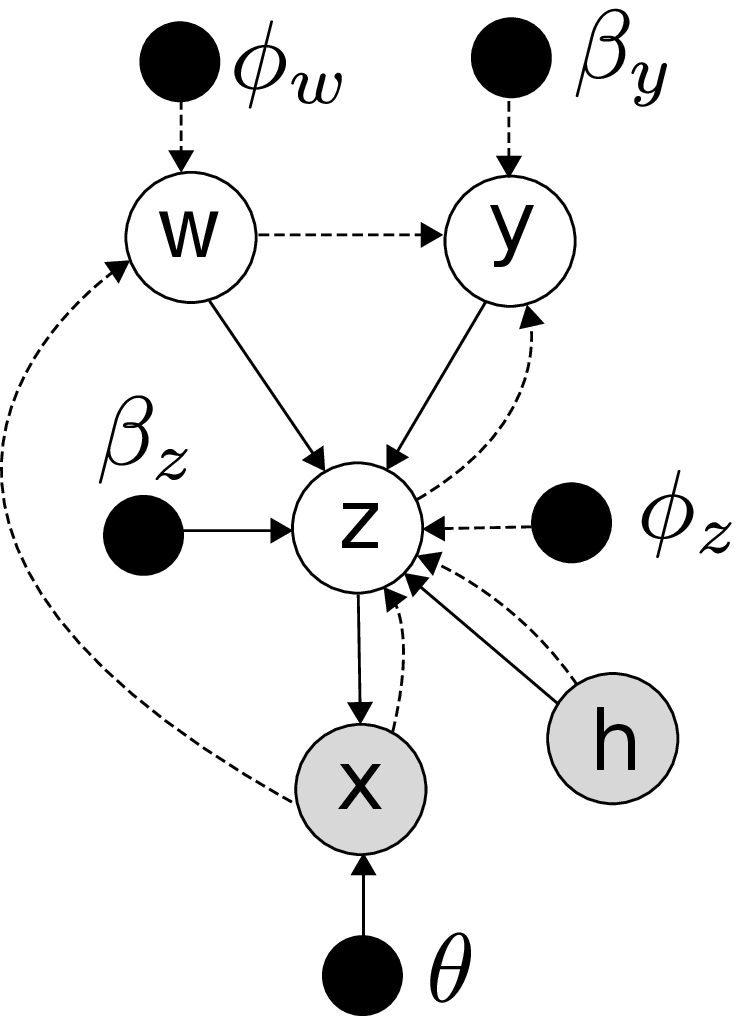}  
		\caption{C-GMVAE model B.}
		\label{fig:graph_cgmvaeB}
	\end{subfigure}
	\caption{The directed graphical models considered for the C-GMVAE in the work. Solid lines denote the generative model and dashed lines the variational approximation.}
	\label{fig:graph_cgmvae}
\end{figure}

The first architecture (model A) that we tried is shown in Figure \ref{fig:graph_cgmvaeA}. For its implementation we had to change the prior distribution of $w$ as $p(w|h) \sim \mathcal{N} (\mu (h), \Sigma (h))$, where the mean and variance of the normal distribution are parameterized by dense nets, and $q_{{\phi}_w} (w|x, h) =\mathcal{N}(\mu_{{\phi}_w}([x; h]), \Sigma_{\phi_w}([x; h]))$, where we only concatenate $h$ to the original input $x$. The main drawback of this model is that the reconstruction as we performed it (compute $z$ from $x$ through the inference model and then reconstruct $x$ from this $z$ by the generative model) does not use the observed $h$.

The graph in Figure \ref{fig:graph_cgmvaeB} belongs to our second version (model B), the one we applied to the presented results. In contrast, in the generative model, we now maintain the original prior of $w$ but condition the $z$ distribution on $h$ as Equation \ref{eq:pz_wyh}. For the variational family, we modify the encoding of $z$ as Equation \ref{eq:pz_xh}.

\begin{subequations}
	\begin{equation}
		\label{eq:pz_wyh}
		p_{\beta_z} (z |w, y, h) = 
		\prod_{k=1}^{K} \mathcal{N} ( \mu_{\beta_z}([w; h]), \Sigma_{\beta_z}([w; h]))^{y_k == 1}
	\end{equation}
	
	\begin{equation}
		\label{eq:pz_xh}
		q_{\phi_z} (z |x, h) = 
		\prod_{k=1}^{K} \mathcal{N} ( \mu_{\phi_z}([x; h]), \Sigma_{\phi_z}([x; h]))
	\end{equation}
\end{subequations}

\section{Dataset for the extension results}
\label{sec:appendixDataset}

In Sections \ref{subsec:exp_seq} and \ref{subsec:exp_seqAtt} we train the models with the \textbf{multi30k} dataset \citep{elliott2016multi30k}. It consists on a training set of $29000$ English sentences and a set of $1000$ test sentences. The vocabulary size is $10118$ tokens. It is not a very large corpora, but we mask some tokens so the scenario gets more complicated to be trained. We use different strategies in the masking process, with higher and lower rates.

In the first strategy, we use a policy of masked tokens more sophisticated that permit the masking of a less percentage of words but focusing on nouns, verbs, adjectives\ldots \quad That is, ignoring stopwords. We use the English stopwords list from the \textit{nltk}\footnote{https://www.nltk.org/}  library \citep{loper2002nltk}. We mask the $80\%$ of the sentences and  we generate two masks in a sentence with a probability of $0.8$. Among these, we also generate a third mask with probability of $0.8$. 

In the second strategy, we increase the number of masked tokens and do not exclude any type of grammatical word so any one can be deleted. In this policy, we mask each token with a probability of $0.6$. Therefore, we have more [MASK] tokens than proper words.

\section{Configuration and experiments}
\label{sec:appendixConf}

\subsection{FMNIST}
\label{subsec:appendixConf_fmnist}
The model used for the experiments with the FMNIST dataset consists on 9 linear layers with RELU as the activation function. The size of the output features on each layer is, from bottom to top, $700$, $600$, $512$, $256$, $128$, $64$, $32$, $16$ and $10$, which corresponds with the number of classes. We employ a negative log-likelihood loss function and the stochastic gradient descent with a learning rate of $0.01$ for the optimization.

The dataset is composed of 28x28 images in grayscale associated with a label from $10$ classes. We divide the set in $12000$ samples for training and $48000$ for validation. The test set has $10000$ images. The only preprocessing step is the normalization to $0.5$ mean and variance.

\subsection{Seq2seq}
\label{subsec:appendixConf_seq2seq}
For this model, we follow the networks structure from \citet{luong2015effective}, omitting the attention mechanism for now. Consequently, we will use a LSTM as the RNN unit, a bidirectional encoder, a depth of two layers in the networks and a hidden size of $1024$ for each of them. 

The configuration of this model follows a seq2seq pre-training of $120$ epochs and a fine-tuning of the regularized decoder for only $20$ epochs after training the GMVAE. In the GMVAE, after different proves we finally chose $1500$ for the hidden dimension, $100$ for $z$, $20$ for $w$ and a $K$ of $10$ MoG of the prior. The depth in the networks is 5 layers and the deviation, $\sigma$, of the posterior normal distribution in the decoder ${10}^{-4}$. We saved the hidden states (encoder output) of all the training sentences, and trained the GMVAE for $100$ epochs.

\subsection{Seq2seq with attention}
\label{subsec:appendixConf_seq2seqAtt}
We keep the same configuration from \citet{luong2015effective}, but including the global attention mechanism.

In the option 1, the C-GMVAE is trained with consecutive pairs of hidden states from the whole training set during $100$ epochs. We finally used a hidden dimension of $1500$, $150$ for the latent space of $z$ and $50$ for $w$. We configured $K=20$ classes in the MoG and a  $\sigma$ of ${10}^{-2}$ for the decoder posterior. The number of layers on each of the distributions modeled was $6$. During the training we selected a learning rate of ${10}^{-5}$, a dropout of $0.3$ and a batch size of $64$. 

In the option 2, the GMVAE is trained for $150$ epochs with the same configuration as before. It only changes the graph as described in Section \ref{subsec:vae}. 

In both options, for the pre-training of the seq2seq, $30$ epochs  were enough since the attention mechanism eases the convergence of the model. After the training of the C-GMVAE and the GMVAE respectively, we fine-tuned the seq2seq decoder with the inclusion of the suitable stochastic layer as mentioned in Section \ref{subsec:method_seqAtt} during other $30$ epochs.

\section{Other models}
\label{sec:appendixOther}

\subsection{Seq2seq with attention}
In the autoregressive model of seq2seq with attention, we, firstly, tried training the C-GMVAE from Figure \ref{fig:graph_cgmvaeA} to generate each hidden state conditioned on the previous one. These generated states were the inputs for the next LSTM unit. However, it did not work as good as we expected. Consequently, we changed the process in a way that instead of using the DGM to generate samples, we could take advantage of its latent space and reconstruct the original hidden states from the LSTMs. 

Inspired by the first model (Section \ref{subsec:method_seq2seq}), we also tried to follow the same idea that is presented as option 1 in Section \ref{subsec:method_seqAtt} but conditioning always in the encoder output instead of the previous hidden state, but it did not improve neither the results that we are presenting in this work. 

Regarding the option 2, initially we used a simpler approach, regularizing the attention output after concatenating it with the LSTM output and exactly before applying the classification layer that matches the vocabulary size. Here, the training was not successful and the imputed words did not follow grammatical rules as a LM is expected to do. After this, we tried the integration of the GMVAE in a previous step, as it is successfully explained in this work.

\section{Additional results}
\label{sec:appendixE}

Table \ref{table:seq2seqAttResults2} presents additional results from the option 2 in the seq2seq with attention model using the \textit{snli} dataset. Once again, we prove the efficacy of our method, even if the dataset gets more complicated. In this table we present different samples of sentences reconstructed from the masked template, following the same philosophy of the results in Table \ref{table:seq2seqAttResults}. The fifth example exposes an extreme case where it is only observed the first word, `a', and both the baseline and our method infer completely different sequences but good alternatives at the same time.

Table \ref{table:topNoRBERTResults2} is an extension of Table \ref{table:topNoRBERTResults} with results form Top NoRBERT.

\begin{table*}[th]
	\centering
	\begin{tabular}{l}
		\toprule
		an \underline{old} man with a package poses \underline{in} front of \underline{an} advertisement .  \\
		an \underline{old} man \textcolor{red}{is standing with arms} \underline{in} front of \underline{an} \textcolor{red}{audience} .  \\
		an \underline{old} man \textcolor{red}{in a blue shirt} \underline{in} front of \underline{an} \textcolor{red}{audience} . \\
		\midrule
		\underline{a man} playing \underline{an} electric \underline{guitar} on stage .\\
		\underline{a man} playing \underline{an} electric \underline{guitar} on stage .  \\
		\underline{a man} \textcolor{red}{plays} \underline{an} electric \underline{guitar} \textcolor{red}{and sings }.  \\
		\midrule
		a blond-haired doctor and her african \underline{american assistant} looking threw \\ \hspace{1cm} new \underline{medical manuals} . \\
		\textcolor{red}{a man is standing in an} \underline{american assistant} \textcolor{red}{, using a} \underline{medical} \textcolor{red}{apparatus .} \\
		\textcolor{red}{a man is looking at the} \underline{american} \textcolor{red}{nurse to get a} \underline{medical} \textcolor{red}{patient .}  \\
		\midrule
		\underline{a young} \textcolor{red}{family enjoys} \underline{feeling ocean} \textcolor{red}{waves lap at their feet} . \\
		\underline{a young} \textcolor{red}{boy is} \underline{feeling ocean} \textcolor{red}{and is on the beach} .  \\
		\underline{a young} \textcolor{red}{man in} \underline{feeling ocean} \textcolor{red}{is surfing on a surfboard} . \\
		\midrule
		\underline{a} \textcolor{red}{man reads the paper in a bar with green lighting .}  \\
		\underline{a} man \textcolor{red}{is standing in front of a crowd of people .}  \\
		\underline{a} man \textcolor{red}{is sitting on a bench reading a book while sitting} \\
		\midrule
		\underline{three} firefighter \underline{come} out of subway station . \\
		\underline{three} \textcolor{red}{people} \underline{come} \textcolor{red}{down a street corner} .\\
		\underline{three} \textcolor{red}{people} \underline{come} out of \textcolor{red}{a boat} . \\
		\midrule
		\underline{a} person \underline{wearing} a \underline{straw} hat \underline{, standing outside} working a \underline{steel} apparatus \\ \hspace{1cm} \underline{with} a pile of coconuts on \underline{the ground .}\\
		\underline{a} \textcolor{red}{man} \underline{wearing} a \underline{straw} hat \underline{, standing outside} \textcolor{red}{of} a \underline{steel} \textcolor{red}{structure} \underline{with} a \\ \hspace{1cm} \textcolor{red}{blue umbrella laying} on \underline{the ground .}\\
		\underline{a} \textcolor{red}{man} \underline{wearing} a \underline{straw} hat \underline{, standing outside} \textcolor{red}{a large} \underline{steel} \textcolor{red}{structure} \underline{with} a \\ \hspace{1cm} \textcolor{red}{tree in front of} \underline{the ground .} \\
		\bottomrule
	\end{tabular}
	\caption{Additional examples of sentences reconstructed by the regularized hidden states in the seq2seq with attention. Sentences order: original, baseline and reconstruction from our regularized option 2 of the GMVAE and the context vectors.}
	\label{table:seq2seqAttResults2}
\end{table*}

\begin{table*}[h]
	\centering
	\begin{tabular}{l}
		\toprule
		\underline{A man looking over a bicycle 's rear wheel in the maintenance garage with} \\ \hspace{1cm} various \underline{tools visible in the background . }\\
		a man looking over a bicycle's \textcolor{red}{back} wheel in the maintenance garage with \\ \hspace{1cm} \textcolor{red}{his} tools visible in the background. \\
		a man looking over a bicycle's rear wheel in \textcolor{red}{a construction} garage with \\ \hspace{1cm} \textcolor{red}{wooden equipment is} in the background.  \\
		\midrule
		\underline{A person} dressed \underline{in a dress with flowers and a stuffed} bee \underline{attached to it ,} \\ \hspace{1cm} \underline{is pushing a baby stroller down the street .} \\
		a person dressed in a \textcolor{red}{suit} with flowers and a stuffed \textcolor{red}{animal} attached to it,\\ \hspace{1cm} is pushing a baby stroller down the street. \\
		a person dressed in a \textcolor{red}{shirt} with flowers and \textcolor{red}{pink} stuffed \textcolor{red}{toy over} to it, is \\ \hspace{1cm} \textcolor{red}{riding} a baby stroller down the street.  \\
		\midrule
		\underline{A blond-haired doctor and her African american assistant looking threw} \\ \hspace{1cm} \underline{new medical manuals .}\\
		a blond - haired doctor and her african american \textcolor{red}{doctor} looking \textcolor{red}{at} new \\ \hspace{1cm} medical \textcolor{red}{scrubs}.\\
		a blond - haired \textcolor{red}{nurse} and her african \textcolor{red}{asian owner} looking \textcolor{red}{around} new \\ \hspace{1cm} medical \textcolor{red}{equipments}. \\
		\midrule
		\underline{3 young man in hoods standing in the middle of a quiet street facing the} \\ \hspace{1cm} \underline{camera .}\\
		3 young man in hoods standing in the middle of a \textcolor{red}{busy} street facing the \\ \hspace{1cm} camera.\\
		a young man in \textcolor{red}{sunglassess} standing in the \textcolor{red}{front} of a \textcolor{red}{busy} street \textcolor{red}{holding} \\ \hspace{1cm} the camera. \\
		\bottomrule
	\end{tabular}
	\caption{Additional examples of sentences reconstructed by Top NoRBERT. The first sentence is the original one, with the observed words underlined. The second is the output of the baseline BERT fine-tuned. Finally, we show the reconstruction. The words in red correspond to mismatches with the original sentence.}
	\label{table:topNoRBERTResults2}
\end{table*}

\end{document}